\documentclass[lettersize,journal]{IEEEtran}
\usepackage{amsmath,amsfonts}
\usepackage{algorithmic}
\usepackage{algorithm}
\usepackage{array}
\usepackage[caption=false,font=normalsize,labelfont=sf,textfont=sf]{subfig}
\usepackage{textcomp}
\usepackage{stfloats}
\usepackage{url}
\usepackage{verbatim}
\usepackage{graphicx}
\usepackage{cite}
\usepackage{amsmath,amssymb,amsfonts}
\usepackage{xcolor}
\usepackage{makecell}
\usepackage{colortbl}
\usepackage{multirow}
\usepackage{bbding}
\usepackage{hyperref}
\usepackage{mathrsfs}
\usepackage{orcidlink}

\begin{document}
\definecolor{mygray}{gray}{.9}
\title{CDXLSTM: Boosting Remote Sensing Change Detection with Extended Long Short-Term Memory}


\author{Zhenkai Wu~\orcidlink{0009-0000-0613-0584}, Xiaowen Ma~\orcidlink{0000-0001-5031-2641}, Rongrong Lian~\orcidlink{0009-0005-3262-6651}, Kai Zheng~\orcidlink{0009-0005-1839-476X}, and Wei Zhang*~\orcidlink{0000-0002-4424-079X}
\thanks{This work was supported by the Public Welfare Science and Technology Plan of Ningbo city (2022S125) and the Science and Technology Innovation 2025 Major Project of Ningbo City (Grant No. 2022Z032).}
\thanks{Zhenkai Wu, Xiaowen Ma, Rongrong Lian, and Kai Zheng are with the School of
Software Technology, Zhejiang University, Hangzhou 310027, China.}
\thanks{Wei Zhang is with the School of Software Technology, Zhejiang University, Hangzhou 310027, China, and also with the Innovation Center of Yangtze River Delta, Zhejiang University, Jiaxing, Zhejiang 314103, China.}
\thanks{\emph{*Corresponding author: Wei Zhang}, e-mail: cstzhangwei@zju.edu.cn.}
}

\markboth{IEEE GEOSCIENCE AND REMOTE SENSING LETTERS,~Vol.~XX, No.~X, XXX~2024}%
{Shell \MakeLowercase{\textit{et al.}}: A Sample Article Using IEEEtran.cls for IEEE Journals}

\IEEEpubid{\begin{minipage}{\textwidth}\ \centering
		XXXX-XXXX~\copyright~2024 IEEE. Personal use is permitted, but republication/redistribution requires IEEE permission. \\
		See https://www.ieee.org/publications/rights/index.html for more information.
\end{minipage}}

\maketitle

\begin{abstract}
In complex scenes and varied conditions, effectively integrating spatial-temporal context is crucial for accurately identifying changes. However, current RS-CD methods lack a balanced consideration of performance and efficiency. 
CNNs lack global context, Transformers are computationally expensive, and Mambas face CUDA dependence and local correlation loss.
In this paper, we propose CDXLSTM, with a core component that is a powerful XLSTM-based feature enhancement layer, integrating the advantages of linear computational complexity, global context perception, and strong interpret-ability.
Specifically, we introduce a scale-specific Feature Enhancer layer, incorporating a Cross-Temporal Global Perceptron customized for semantic-accurate deep features, and a Cross-Temporal Spatial Refiner customized for detail-rich shallow features.
Additionally, we propose a Cross-Scale Interactive Fusion module to progressively interact global change representations with spatial responses. Extensive experimental results demonstrate that CDXLSTM achieves state-of-the-art performance across three benchmark datasets, offering a compelling balance between efficiency and accuracy.
Code is available at https://github.com/xwmaxwma/rschange.
\end{abstract}

\begin{IEEEkeywords}
Remote Sensing Change Detection, Extended Long Short-Term Memory, Spatio-Temporal Interaction.
\end{IEEEkeywords}

\section{Introduction}

\IEEEPARstart{T}{he} advancement of Earth observation technology, including better remote sensing platforms and sensors, has improved the ability to monitor surface activities. Remote Sensing Change Detection (RS-CD) identifies changes by comparing images over time, aiding urban planning \cite{marin2014building}, disaster assessment \cite{mahdavi2019polsar}, and environmental monitoring \cite{de2020change}.

RS-CD tasks are inherently multi-scale and multi-temporal, with effective change detection relying on the aggregation of spatial and temporal context. CNN-based methods introduced deep learning to RS-CD by designing multi-scale feature fusion structures for improved spatio-temporal modeling \cite{steinformer}. Techniques such as deeper CNNs \cite{zhang2020feature}, dilated convolutions \cite{zhang2018triplet}, attention mechanisms \cite{peng2020optical}, multi-scale convolutions \cite{USSFCNet}, and the inner fusion properties of 3D convolutions \cite{AFCF3DNet} have been extensively explored. Moreover, HFIFNet \cite{han2025hfifnet} enhances spatiotemporal information interaction by incorporating the dual-branch interaction strategy, which captures spatial correlations between bi-temporal images and improves changed target identification. However, challenges persist in effectively modeling long-range dependencies. After that, Transformer-based methods have gained traction for RS-CD due to their global self-attention mechanism, modeling spatio-temporal dependencies \cite{changeformer, BIT}. These approaches focus on cross-scale \cite{cdmask} and cross-temporal fusion \cite{SARASNet}, achieving strong results in global spatio-temporal modeling. However, they often suffer from quadratic computational complexity caused by the self-attention calculation. Although some linear Transformers effectively address computational complexity \cite{aff, flatten}, they are often at the cost of sacrificing the global reception.

Recently, Mamba-based methods \cite{rsmamba, wu2025cd} have gained rapid popularity for their linear complexity and global perception capabilities. However, their reliance on CUDA depencence (i.e., the underlying CUDA operators need to be restructured, and the data interaction mechanisms redesigned to fully leverage GPUs for efficient parallel computation \cite{gu2023mamba}.) and suboptimal performance remain limitations. Their sequential scanning may inadvertently disrupt the strong local correlations inherent in image structures. In response, XLSTM introduces an exponential gating mechanism and matrix-parallel memory \cite{beck2024xlstm, xlstm}, combining enhanced interpretability through dynamic memory updates based on current and historical data, and parallel acceleration via matrix memory. We aim to apply XLSTM to RS-CD for the first time, enabling more intuitive and efficient change representation capture.

\begin{figure*}[h]
  \centering
  \includegraphics[width=0.88\linewidth]{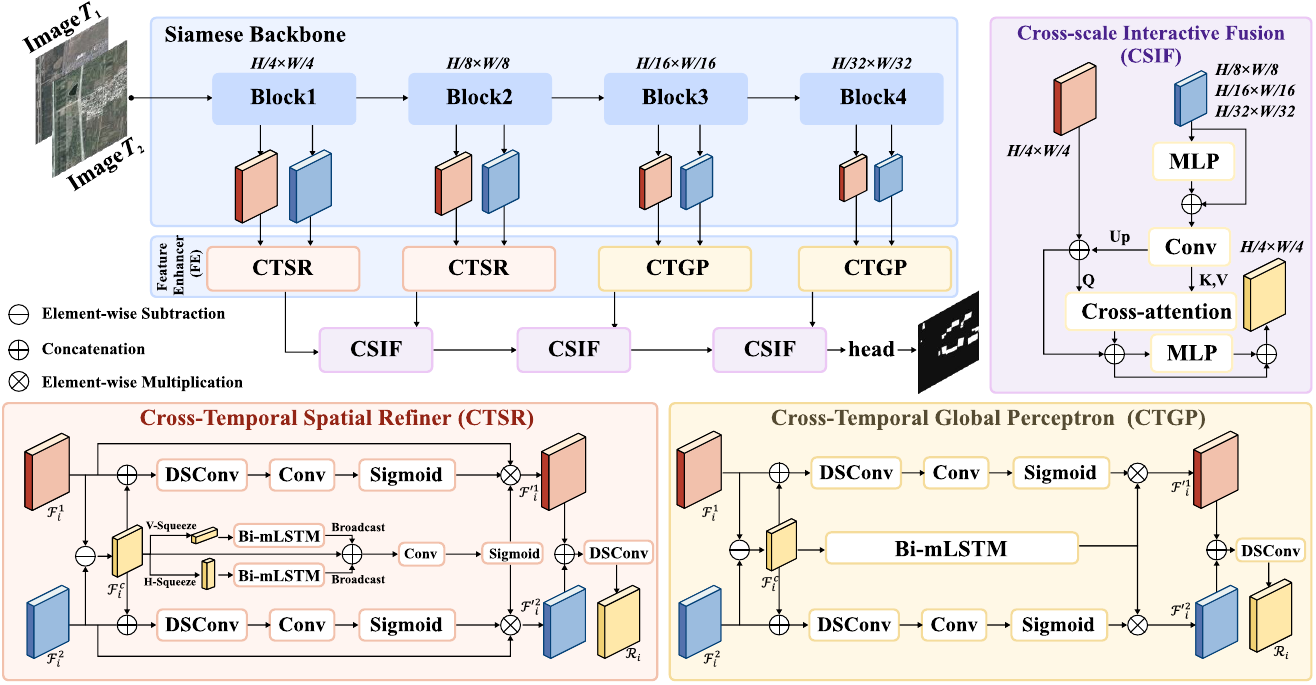}
  \caption{CDXLSTM architecture takes a pair of co-registered remote sensing images as input into a Siamese Backbone with shared weights, producing two feature maps at each stage. CTSR is applied in the first two shallow stages to refine spatial changes, while CTGP is used in the latter two deep stages to capture global changes. Here, Bi-mLSTM refers to an mLSTM module that performs bidirectional scanning. These feature maps are then progressively aggregated through the CSIF module.}
  \label{fig:net}
\vspace{-3mm}
\end{figure*}


\IEEEpubidadjcol
In this paper, we evaluate the strengths and limitations of CNNs, Transformers, and Mambas, while highlighting the potential of XLSTM for RS-CD tasks. Bi-temporal images generate multi-scale feature maps through a Siamese backbone. 
The CDXLSTM’s core XLSTM-based feature enhancement (FE) layer addresses key challenges such as local information loss during scanning and shallow feature inaccuracy. Specifically, the Cross-Temporal Global Perceptron (CTGP) employs XLSTM to extract global context from deep feature maps, enhancing semantic differences. Meanwhile, the Cross-Temporal Spatial Refiner (CTSR) integrates XLSTM with axial attention to reduce noise, refine spatial details, and address shallow layer limitations.
Specifically, we employed a shared-parameter bidirectional scanning mLSTM block \cite{xlstm} (Bi-mLSTM). Finally, we recognize that the largest-scale branch, with its comprehensive spatial information, is crucial for accurate change detection. We propose a Cross-scale Interactive Fusion module (CSIF), which uses the largest-scale branch as a foundation to progressively integrate spatial information and global semantics from smaller-scale branches. Overall, our contributions can be summarized as follows:

\begin{itemize}
    \item[1)] We analyze the potential of XLSTM in comparison to CNNs, Transformers, and Mambas, and applied it to RS-CD tasks for the first time. With its linear complexity, global context awareness, parallel acceleration, and enhanced interpret-ability, XLSTM enables more intuitive and efficient differentiation of changes of interest.
    \item[2)] We design a scale-specific XLSTM scanning strategy tailored for RS-CD tasks to reduce both local correlation loss and global redundancy. Specifically, we introduce the Cross-Temporal Global Perceptron (CTGP) and Cross-Temporal Spatial Refiner (CTSR) to effectively capture both fine spatial details and global contextual changes.
    \item[3)] We propose a Cross-scale Interactive Fusion module (CSIF) to progressively integrate spatial information and global semantics.
\end{itemize}

\section{Methodology}

\subsection{Overall Architecture}
\label{OA}
We propose CDXLSTM (Fig. \ref{fig:net}), which strikes a good balance between performance and efficiency. It employs a scale-specific XLSTM scanning strategy that effectively addresses differences between scales, further enhancing performance.

Specifically, CDXLSTM includes a Siamese backbone, a FE layer (with CTGP and CTSR), and a CSIF module. Features are extracted using weight-shared Seaformer-L \cite{seaformer} Siamese backbones, with the low-resolution branch capturing global semantics via CTGP and the high-resolution branch enhancing spatial details via CTSR. Finally, the CSIF module enables effective multi-scale feature interaction.

\setlength{\tabcolsep}{3pt}
\begin{table*}[t]
	\begin{center}
		\caption{
		Comparison of performance for RSCD on LEVIR-CD, WHU-CD, and CLCD datasets. Highest scores are in bold. All scores are in percentage. Flops are counted with image size of $256 \times 256 \times 3$.
		}
		\label{table:1}
            \begin{tabular}{l||cc||ccccc||ccccc||ccccc}
		\Xhline{1.2pt}
            \rowcolor{mygray}
		     & & &\multicolumn{5}{c||}{LEVIR-CD} &\multicolumn{5}{c||}{WHU-CD} &\multicolumn{5}{c}{CLCD}\\
            \rowcolor{mygray}
			\multicolumn{1}{c||}{\multirow{-2}{*}{Method}}
               & \multirow{-2}{*}{\makecell{Params\\(M)}}& \multirow{-2}{*}{\makecell{Flops\\(G)}} &F1 &Pre. &Rec. &IoU &OA  &F1 &Pre. &Rec. &IoU &OA  &F1 &Pre. &Rec. &IoU &OA \\		
                \hline \hline
                FC-EF \cite{fc-siam} &  1.10 & 1.55 &83.4 &86.91 &80.17 &71.53 &98.39 &72.01 &77.69 &67.10 &56.26 &92.07 &48.64 &73.34 &36.29 &32.14 &94.30\\
   FC-Siam-Di \cite{fc-siam}& 1.35 & 4.25  &86.31 &89.53 &83.31 &75.92 &98.67 &58.81 &47.33 &77.66 &41.66 &95.63 &44.10 &72.97 &31.60 &28.29 &94.04\\
   FC-Siam-Conc \cite{fc-siam}& 1.55 & 4.86 &83.69 &91.99 &76.77&71.96&98.49 &66.63 &60.88 &73.58 &49.95 &97.04 &54.48&68.21&45.22&37.35&94.35 \\
   IFNet \cite{ifnet}& 50.71  &  41.18  &88.13 &\bf94.02 &82.93&78.77&98.87 &83.40 &\bf96.91 &73.19 &71.52 &98.83 &48.65&49.96&47.41&32.14&92.55\\
   DTCDSCN \cite{dtcdscn}& 41.07  &  14.42 &87.67 &88.53&86.83&78.05&98.77 &71.95 &63.92 &82.30 &56.19 &97.42 &60.13&62.98&57.53&42.99&94.32\\
   BIT \cite{BIT}& 11.89  & 8.71 &89.31 &89.24 &89.37&80.68&98.92 &83.98 &86.64 &81.48 &72.39 &98.75&67.10&73.07&62.04&50.49&95.47 \\
   SNUNet \cite{snunet}&  12.03  & 46.70 &88.16 &89.18 &87.17&78.83&98.82 &83.50 &85.60 &81.49 &71.67 &98.71  &66.70&73.76 &60.88&50.04&95.48\\
   DMATNet \cite{dmatnet} &  13.27 & - & 89.97 &90.78 &89.17&81.83 & 98.06 &85.07 &89.46 &82.24 &74.98 &95.83 &66.56&72.74&61.34&49.87&95.41 \\
   LGPNet \cite{lgpnet}& 70.99  &  125.79 &89.37 &93.07 &85.95&80.78&99.00 &79.75 &89.68 &71.81 &66.33 &98.33&63.03&70.54&56.96&46.01&95.03\\
   ChangeFormer \cite{changeformer}& 41.03 & 202.86 &90.40 &92.05&88.80&82.48&99.04 &81.82 &87.25 &77.03 &69.24 &94.80&58.44&65.00&53.07&41.28&94.38\\
   SARASNet \cite{SARASNet}& 56.89 & 139.9 &  90.44   &91.42  &89.48 &82.55  & \bf99.11  & 89.55 & 88.68 & 90.44  & 81.08 & 99.05  & 74.70 & 76.68 & 72.83 & 59.62  & 96.33\\
   USSFC-Net \cite{USSFCNet} & 1.52 & 3.17 & 88.80 & 87.18 & 90.49 & 79.86 & 98.84 & 88.93 & 91.56 & 86.43 & 80.06 & 99.01 & 63.04 & 64.83 & 61.34 & 46.03 &  94.42\\
   AFCF3D-Net \cite{AFCF3DNet} & 17.64 & 31.58 & 90.44 & 91.18 & 89.72& 82.55& 99.03& 92.07& 93.56&90.62 & 85.30& 99.28& 76.92 & \bf84.20 & 70.79 & 62.49 & 96.84\\
   RS-Mamba \cite{rsmamba} & 51.95 & 22.82 & 90.67 & 90.70 & \bf90.63 & 82.93 & 99.05 & 91.50 & 93.21 & 89.85 & 84.33 & 99.23  &  71.27  &  72.95 &  69.67 & \bf72.95 & 95.82\\
   ChangeMamba \cite{changemamba}&  48.56 & 38.49 & 90.51 & 90.73 & 90.30 & 82.67& 99.04 & 90.08 & 92.94 & 87.38 & 81.95 & 99.12 & 70.00 & 75.39 & 65.33 & 53.85 & 95.83\\
   			\hline
   Ours & 16.19 & 3.92 &  \bf90.89& 91.52 & 90.27 & \bf83.30 & 99.07 & \bf92.58 & 93.71 & \bf91.49 & \bf86.19 & \bf99.33& \bf78.73 & 83.15 & \bf74.76  & 64.92 & \bf96.99\\
			\hline
		\end{tabular}
  \vspace{-3mm}
	\end{center}
\end{table*}

\subsection{Feature Enhancer}
\label{FE}
 We believe the low-resolution branch, with its global perspective, should primarily be used to differentiate between change regions and background in bi-temporal features. In contrast, the high-resolution branch, rich in spatial details, should focus on enhancing spatial responses, which is crucial for accurately locating change regions. Therefore, we leverage XLSTM's global perception capability to design the CTGP for the low-resolution branch, while utilizing its long-term modeling ability to  refine local details by CTSR for the high-resolution branch, each tailored to their specific roles.

\subsubsection{Cross-Temporal Global Perceptron}
\label{CTGP}
This block captures global change semantics for the low-resolution branch. Given the low-resolution bi-temporal features $F_{i}^{1}$ and $F_{i}^{2}$ extracted by Siamese backbones, we first compute the coarse global change representation $F_{i}^{c}$ via element-wise subtraction:

\begin{equation}
\label{eqn1}
F_{i}^{c}=F_{i}^{1}\ominus F_{i}^{2}
\end{equation}

Next, $F_{i}^{c}$ is concatenated with each of the bi-temporal features separately. This is followed by a deep separable convolution (DSConv) and a sigmoid activation function ($\sigma$), enabling the model to learn finer bi-temporal global semantic details and generate more accurate attention weights, resulting in $W_{i}^{1}$ and $W_{i}^{2}$, as shown in Eq. \ref{eqn2}. 

\begin{equation}
\label{eqn2}
W_{i}^{t}=\sigma \left( \mathrm{Conv}\left( \mathrm{DSConv}\left( F_{i}^{c}\oplus F_{i}^{t} \right) \right) \right) ,t\in \left\{ 1,2 \right\} 
\end{equation}

We then introduce the mLSTM module, the key parallelizable long-term perception module of XLSTM, to directly apply long-term modeling to $F_{i}^{c}$ and enhance the global change representation. Since the images are unordered, we use a bidirectional mLSTM (Bi-mLSTM, denoted by $\varPhi$) with shared parameters, including matrix memory, causal convolutions, and multi-head linear layers. The settings are based on \cite{xlstm} to balance interaction and efficiency, where the kernel size of the causal convolution is set to 4, and the number of heads in the mLSTM cells is set to 4. Finally, the spatial weights are used to enhance the change representations, which are then concatenated and passed through another DSConv, resulting in the final global change representation $R_i$. This operation is shown in Eqs. \ref{eqn3} and \ref{eqn4}.

\begin{equation}
\label{eqn3}
{F^{'}}_{i}^{t}=W_{i}^{t}\otimes \varPhi \left( F_{i}^{c} \right) ,t\in \left\{ 1,2 \right\}
\end{equation}

\begin{equation}
\label{eqn4}
R_i=\mathrm{DSConv}\left( {F^{'}}_{i}^{1}\oplus {F^{'}}_{i}^{2} \right) 
\end{equation}

\setlength{\tabcolsep}{2pt}
\begin{table}[t]
	\begin{center}
		\caption{
		Ablation experiences of the long-term modeling strategy on CLCD dataset.
		}
		\label{table:3}
            \begin{tabular}{l||cc||ccccc}
		\Xhline{1.2pt}
            \rowcolor{mygray}
		     Strategy & Parms(M) & Flops(G)  &F1 &Pre. &Rec. &IoU  & OA\\		
                \hline \hline
                CNN & 15.27 & 3.61 & 76.70 &  82.20 & 71.90  & 62.21  & 96.75 \\
                Transformer & 19.25 & 4.01 & 75.50  & 76.69  & 74.34  & 60.64  & 96.41\\
                Mamba & 16.61 & 3.74 & 77.76 & 82.46 &  73.58 &  63.62 &  96.87  \\
                Linear-TR1 & 16.25& 4.42& 77.22& 77.92& \bf76.54& 62.90&  96.64 \\ 
                Linear-TR2 & 18.28 & 5.95 & 77.08 & 82.40 & 72.40 & 62.70 & 96.80\\ 
                DBIM(HFIFNet\cite{han2025hfifnet}) & 16.91 & 4.22 & 77.65 & \bf83.37 & 72.65 & 63.46 & 96.89 \\
                \hline
                Bi-mLSTM & 16.19& 3.92 & \bf78.73 & 83.15 & 74.76  & \bf64.92 & \bf96.99 \\            
   			\hline
		\end{tabular}
	\end{center}
  \vspace{-7mm}
\end{table}

\subsubsection{Cross-Temporal Spatial Refiner}
\label{CTSR}
This block is designed for the high-resolution branch to obtain a change representation with rich spatial information. Given the high-resolution bi-temporal features $F_{i}^{1}$ and $F_{i}^{2}$ extracted by the Siamese backbones, we first obtain the coarse change representation $F_{i}^{c}$ and spatial detail enhancement weights ($W_{i}^{1}$ and $W_{i}^{2}$) according to Eqs. \ref{eqn1} and \ref{eqn2}.

Unlike CTGP, we employ axial Bi-mLSTM as an attention mechanism to model spatial details in $F_{i}^{c}$, particularly suited for changes of interest that predominantly occur in elongated, strip-shaped buildings. Specifically, we first apply average pooling to $F_{i}^{c}$ along the horizontal ($V\_s$) and vertical ($H\_s$) axes, followed by a Bi-mLSTM ($\varPhi$) module for each direction. The outputs are then broadcast back to the original size and combined with $F_{i}^{c}$ via element-wise addition. After passing through a sigmoid activation function ($\sigma$), we obtain the axial spatial enhancement weights ($W_{i}^{c}$), as shown in Eq. \ref{eqn5}.

\begin{equation}
\label{eqn5}
W_{i}^{c}=\sigma \left( \mathrm{Conv}\left( F_{i}^{c}+\varPhi \left( V\_s\left( F_{i}^{c} \right) \right) +
\varPhi \left( H\_s(F_{i}^{c}) \right) \right) \right)
\end{equation}

To preserve as much original spatial detail as possible, both the axial spatial weights and spatial detail enhancement weights are used to refine the original bi-temporal feature map. Finally, the enhanced bi-temporal feature maps are concatenated and processed by DSConv to produce the final change representation $R_i$, as shown in Eq. \ref{eqn6}.

\begin{equation}
\label{eqn6}
R_i=\mathrm{DSConv}\left( \left( W_{i}^{c}\otimes W_{i}^{1}\otimes F_{i}^{1} \right) \oplus \left( W_{i}^{c}\otimes W_{i}^{2}\otimes F_{i}^{2} \right) \right) 
\end{equation}

\vspace{-3mm}
\subsection{Cross-scale Interactive Fusion Module}
\label{CSTF}
After obtaining the scale-specific change representations, we propose the CSIF module to combine spatial details with high-level global change semantics. To accurately pinpoint regions of interest, the highest-resolution change representation, which retains the richest spatial details, should be used as the reference, ensuring the preservation of as much spatial detail as possible. Consequently, subsequent change representations are progressively fused with this reference using the CSIF module.

In the CSIF module, given a high-resolution change representation $R_h$ and a low-resolution change representation $R_l$, we first enhance $R_l$ using an MLP residual block, which enhances the nonlinear fitting capability to bridge the gap between the high-resolution and low-resolution change representations, resulting in $R_l'$. This is then upsampled and added element-wise to $R_h$, producing $R_h'$. Next, $R_h'$ is used as the $\mathcal{Q}$ and $R_l'$ as the $\mathcal{K}$ and $\mathcal{V}$, with cross-attention applied to fuse the high- and low-resolution information, followed by a residual connection with $R_h'$. To reduce computational complexity, $\mathcal{K}$ convolves $\mathcal{Q}$, and $\mathcal{V}$ then convolves the result. After an additional MLP residual enhancement, the aggregated representation $R_h^o$ is obtained. This $R_h^o$ will serve as the high-resolution change representation for the next CSIF module. 

After aggregating through three CSIF modules, we obtain the final change representation $R^o$. This is then passed through two MLP layers, with the first layer incorporating a DSConv residual connection, resulting in the final change map.

\subsection{Loss Function}
\label{loss}
CDXLSTM employs a combination of binary cross-entropy loss and dice loss~\cite{dice} to supervise the mask. The final loss function is a weighted sum of the mask loss and classification loss, defined as: $\mathcal{L} = \lambda_{ce}\mathcal{L}_{ce} + \lambda_{dice}\mathcal{L}_{dice}$. The weighting parameters $\lambda_{ce}$ and $\lambda_{dice}$ are set to a 1:1 ratio.

\section{Experiments}
\subsection{Datasets and Implementation Details}
We perform experiments on the LEVIR-CD \cite{levir}, WHU-CD \cite{whu}, and CLCD \cite{clcd} datasets. We use the F1-score (F1) with regard to the change category as the main evaluation indices. Please refer to the Supplementary Material for more details.

\setlength{\tabcolsep}{3pt}
\begin{table}[t]
	\centering
		\caption{
    Ablation experiences involving the combination between CTSR ($\bot$) and CTGP ($\circledcirc$) on CLCD dataset.
		}
		\label{table:2}
            \begin{tabular}{cccc||c||ccccc}
		\Xhline{1.2pt}
            \rowcolor{mygray}
		     \multicolumn{4}{c||}{Feature Enhancer}  & Flops & &  & & &\\
       \rowcolor{mygray}
             1/4 & 1/8  &  1/16 &  1/32 & (G) & \multicolumn{1}{c}{\multirow{-2}{*}{F1}} & \multicolumn{1}{c}{\multirow{-2}{*}{Pre.}} & \multicolumn{1}{c}{\multirow{-2}{*}{Rec.}} & \multicolumn{1}{c}{\multirow{-2}{*}{IoU}} & \multicolumn{1}{c}{\multirow{-2}{*}{OA}}\\		
                \hline \hline
                $\circledcirc$ & $\bot$ & $\bot$ & $\bot$ & 12.49 & 77.00 & 80.57  &  73.74 &  62.61 & 96.72 \\
                $\circledcirc$ & $\circledcirc$ & $\bot$ & $\bot$  & 13.76 & 76.93 & 80.27 & 73.85 & 62.50 & 96.70 \\
                $\circledcirc$ & $\circledcirc$ & $\circledcirc$ & $\bot$ &  13.96 & 76.97 & 79.25 & 74.82  & 62.57  & 96.67\\
                $\circledcirc$ & $\circledcirc$ & $\circledcirc$ & $\circledcirc$  & 14.00 & 76.87 & 78.95 & 74.91 & 62.44 & 96.65 \\
                $\bot$ & $\circledcirc$ & $\circledcirc$ & $\circledcirc$  & 5.19 & 78.26 & 80.29  & 76.33  & 64.28 & 96.84 \\
                $\bot$ & $\bot$ & $\circledcirc$  & $\circledcirc$ & 3.92 & \bf78.73 & \bf83.15 & \bf74.76  & \bf64.92 & \bf96.99 \\
                $\bot$ & $\bot$ & $\bot$ & $\circledcirc$  & 3.73 & 77.54 & 83.24 & 72.58  & 63.32 & 96.87  \\
                $\bot$ & $\bot$ & $\bot$ & $\bot$ &  3.68  & 77.74 & 81.36 & 74.43  & 63.58 &  96.83                                     \\
   			\hline
		\end{tabular}
\end{table}

\setlength{\tabcolsep}{4pt}
\begin{table}[t]
	\begin{center}
		\caption{
		Ablation experiences of the proposed blocks on CLCD dataset.
		}
		\label{table:4}
            \begin{tabular}{ccc||ccccc}
		\Xhline{1.2pt}
            \rowcolor{mygray}
		     CTSR & CTGP & CSIF & F1 &Pre. &Rec. &IoU  & OA\\		
                \hline \hline
               \XSolidBrush &  \CheckmarkBold&  \CheckmarkBold  & 78.03 & 78.05  & \bf78.01 &  63.98 & 96.73  \\
                \CheckmarkBold &  \XSolidBrush&  \CheckmarkBold  & 76.16 & 83.02 & 70.35  & 61.50  &  96.72\\
               \CheckmarkBold  & \CheckmarkBold & \XSolidBrush & 76.63 & 79.13 & 74.28  & 62.11  & 96.63    \\
                \hline
              \CheckmarkBold  &  \CheckmarkBold&  \CheckmarkBold  & \bf78.73 & \bf83.15 & 74.76  & \bf64.92 & \bf96.99 \\            
   			\hline
		\end{tabular}
	\end{center}
\vspace*{-0.3cm}
\end{table}

\subsection{Main Results}
We compared our results with state-of-the-art methods across various categories, including convolution-based methods such as FC-EF \cite{fc-siam}, FC-Siam-Di \cite{fc-siam}, FC-Siam-Conc \cite{fc-siam}, IFNet \cite{ifnet}, DTCDSCN \cite{dtcdscn}, SNUNet \cite{snunet}, LGPNet \cite{lgpnet}, USSFC-Net \cite{USSFCNet}, and AFCF3D-Net \cite{AFCF3DNet}; transformer-based methods including DMATNet \cite{dmatnet}, BIT \cite{BIT}, ChangeFormer \cite{changeformer}, and SARASNet \cite{SARASNet}; and Mamba-based methods RS-Mamba \cite{rsmamba} and ChangeMamba \cite{changemamba}. 

\begin{figure}[t]
\vspace*{-0.2cm}
\centering
\captionsetup[subfloat]{labelsep=none,format=plain,labelformat=empty,font=tiny}
\subfloat[$T_1$]{
\begin{minipage}[t]{0.11\linewidth}
\includegraphics[width=1\linewidth]{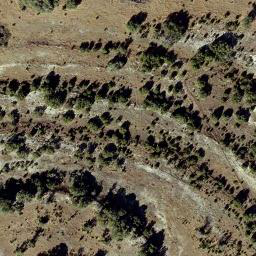}\vspace{2pt}
\includegraphics[width=1\linewidth]{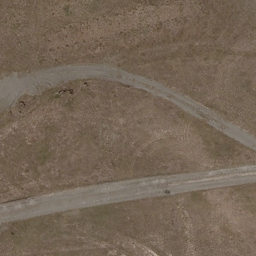}\vspace{2pt}
\includegraphics[width=1\linewidth]{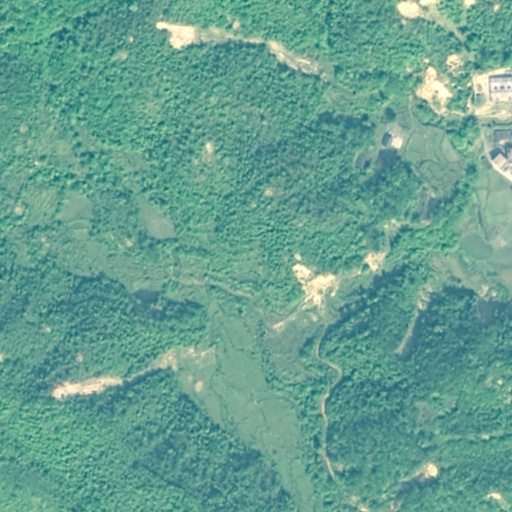}\vspace{2pt}
\end{minipage}}
\subfloat[$T_2$]{
\begin{minipage}[t]{0.11\linewidth}
\includegraphics[width=1\linewidth]{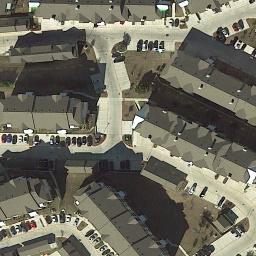}\vspace{2pt}
\includegraphics[width=1\linewidth]{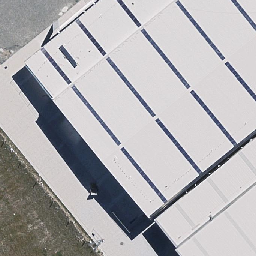}\vspace{2pt}
\includegraphics[width=1\linewidth]{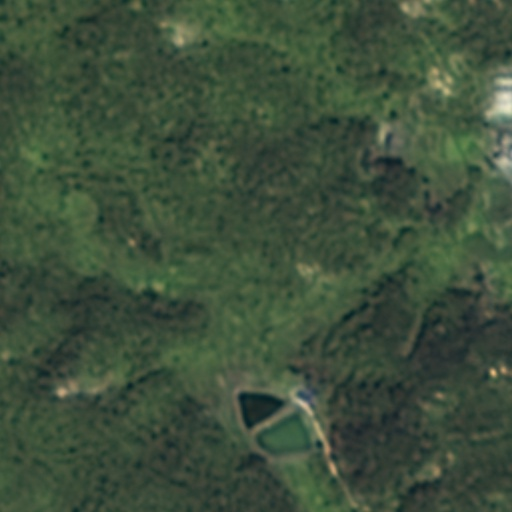}\vspace{2pt}
\end{minipage}}
\subfloat[GT]{
\begin{minipage}[t]{0.11\linewidth}
\includegraphics[width=1\linewidth]{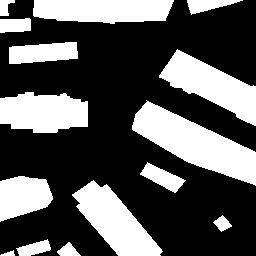}\vspace{2pt}
\includegraphics[width=1\linewidth]{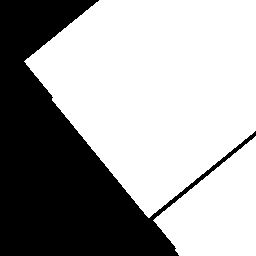}\vspace{2pt}
\includegraphics[width=1\linewidth]{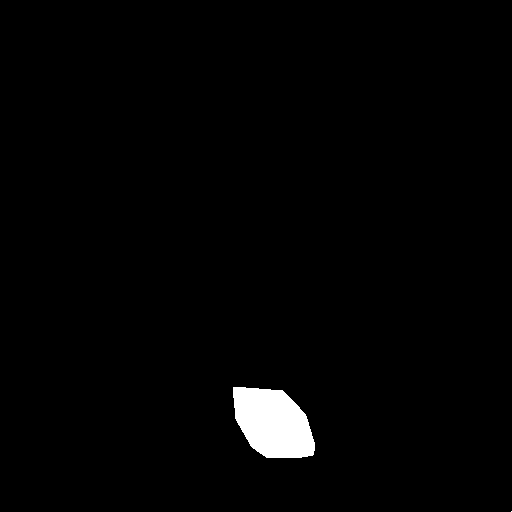}\vspace{2pt}
\end{minipage}}
\subfloat[SNUNet]{
\begin{minipage}[t]{0.11\linewidth}
\includegraphics[width=1\linewidth]{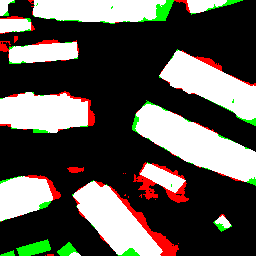}\vspace{2pt}
\includegraphics[width=1\linewidth]{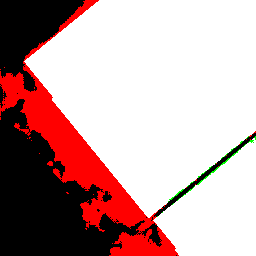}\vspace{2pt}
\includegraphics[width=1\linewidth]{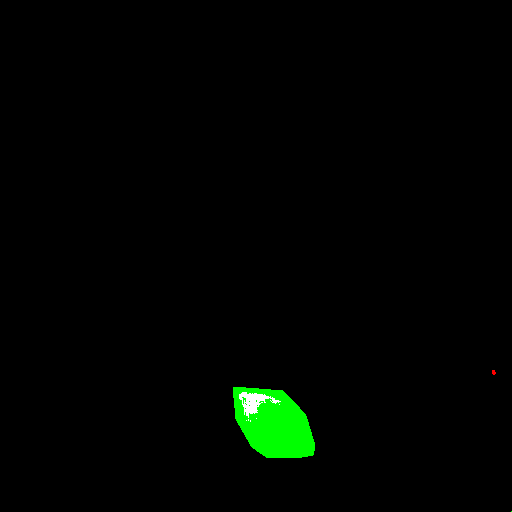}\vspace{2pt}
\end{minipage}}
\subfloat[BIT]{
\begin{minipage}[t]{0.11\linewidth}
\includegraphics[width=1\linewidth]{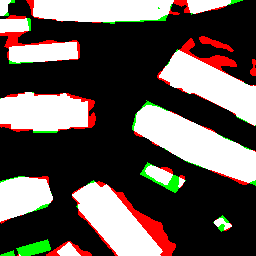}\vspace{2pt}
\includegraphics[width=1\linewidth]{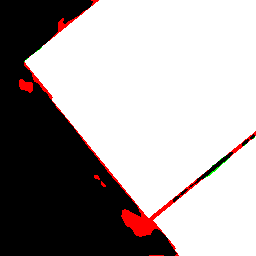}\vspace{2pt}
\includegraphics[width=1\linewidth]{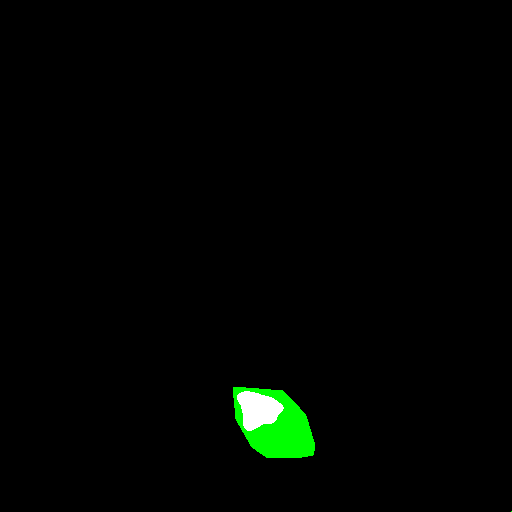}\vspace{2pt}
\end{minipage}}
\subfloat[SARASNet]{
\begin{minipage}[t]{0.11\linewidth}
\includegraphics[width=1\linewidth]{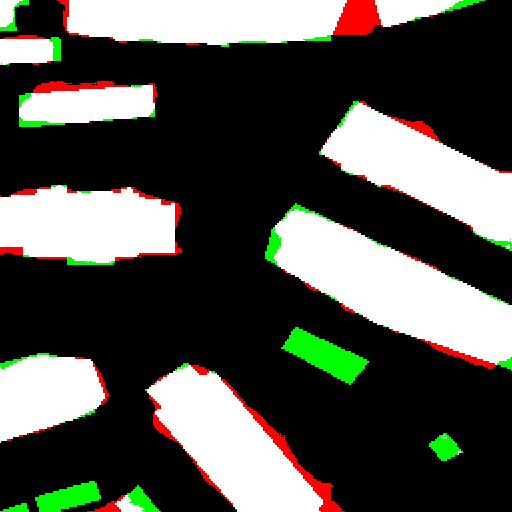}\vspace{2pt}
\includegraphics[width=1\linewidth]{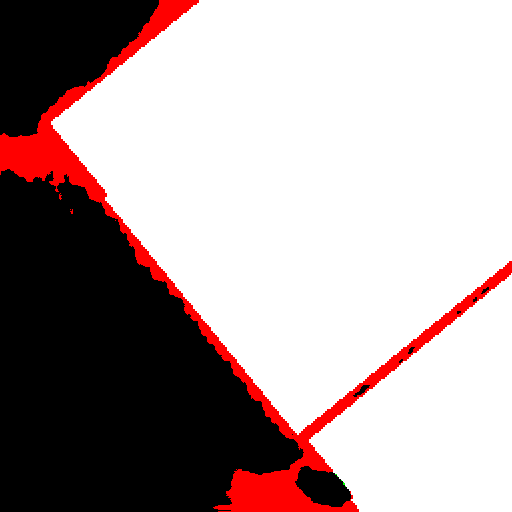}\vspace{2pt}
\includegraphics[width=1\linewidth]{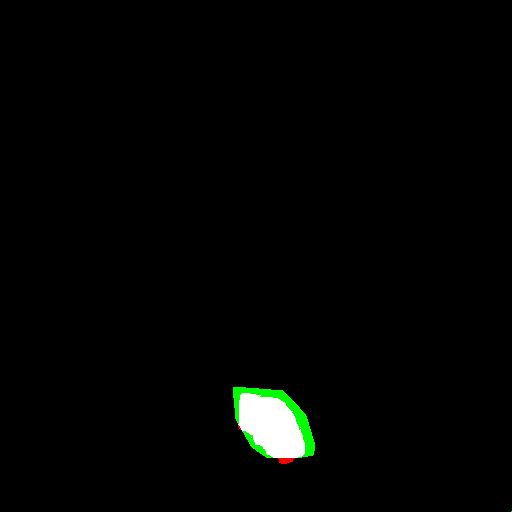}\vspace{2pt}
\end{minipage}}
\subfloat[AFCF3D-Net]{
\begin{minipage}[t]{0.11\linewidth}
\includegraphics[width=1\linewidth]{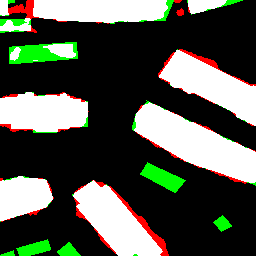}\vspace{2pt}
\includegraphics[width=1\linewidth]{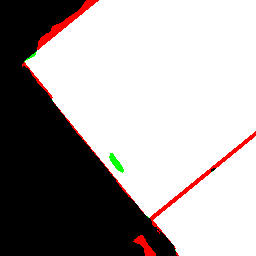}\vspace{2pt}
\includegraphics[width=1\linewidth]{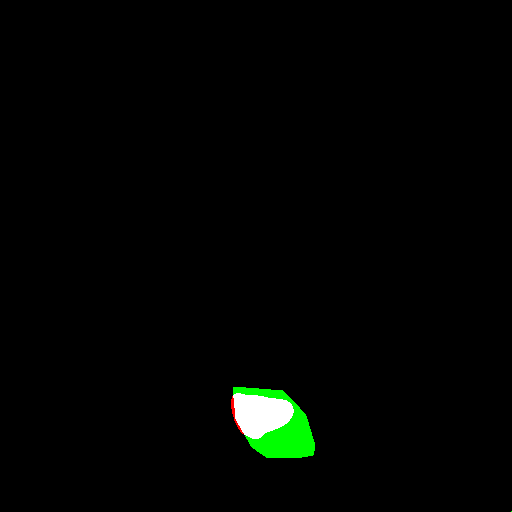}\vspace{2pt}
\end{minipage}}
\subfloat[CDXLSTM]{
\begin{minipage}[t]{0.11\linewidth}
\includegraphics[width=1\linewidth]{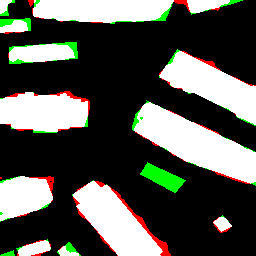}\vspace{2pt}
\includegraphics[width=1\linewidth]{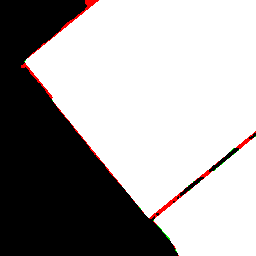}\vspace{2pt}
\includegraphics[width=1\linewidth]{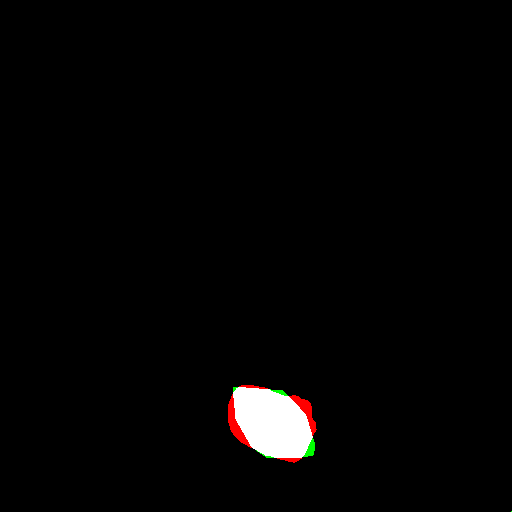}\vspace{2pt}
\end{minipage}}
\caption{Example results on LEVIR-CD (row 1), WHU-CD (row 2), and CLCD (row 3) test sets, with pixel color coding: white for true positives, black for true negatives, red for false positives, and green for false negatives.}
\label{fig:clcd}
\end{figure}

\begin{figure}[t]
\vspace*{-0.5cm}
\centering
\captionsetup[subfloat]{labelsep=none,format=plain,labelformat=empty,font=tiny}
\subfloat[$T_1$]{
\begin{minipage}[t]{0.11\linewidth}
\includegraphics[width=1\linewidth]{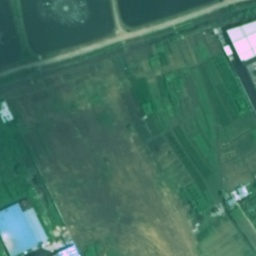}\vspace{2pt}
\includegraphics[width=1\linewidth]{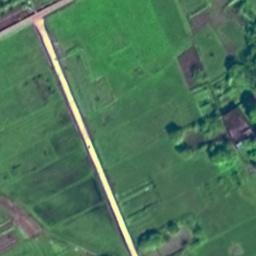}\vspace{2pt}
\end{minipage}}
\hspace{5pt}
\subfloat[$T_2$]{
\begin{minipage}[t]{0.11\linewidth}
\includegraphics[width=1\linewidth]{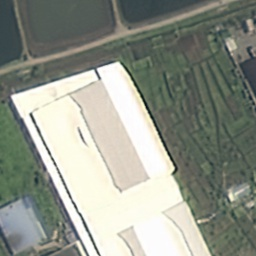}\vspace{2pt}
\includegraphics[width=1\linewidth]{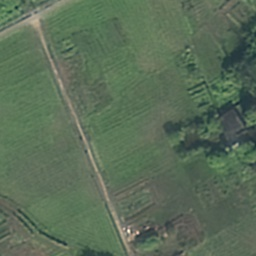}\vspace{2pt}
\end{minipage}}
\hspace{5pt}
\subfloat[GT]{
\begin{minipage}[t]{0.11\linewidth}
\includegraphics[width=1\linewidth]{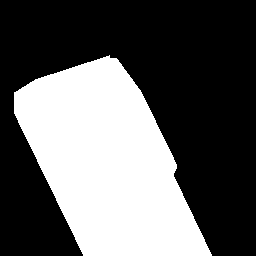}\vspace{2pt}
\includegraphics[width=1\linewidth]{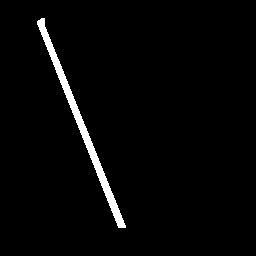}\vspace{2pt}
\end{minipage}}
\hspace{5pt}
\subfloat[$\circledcirc\circledcirc\circledcirc\circledcirc$]{
\begin{minipage}[t]{0.11\linewidth}
\includegraphics[width=1\linewidth]{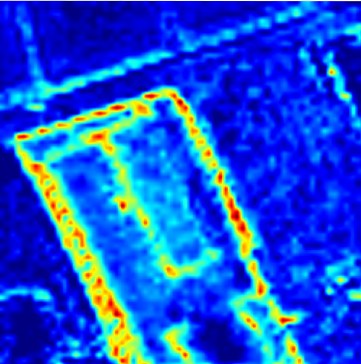}\vspace{2pt}
\includegraphics[width=1\linewidth]{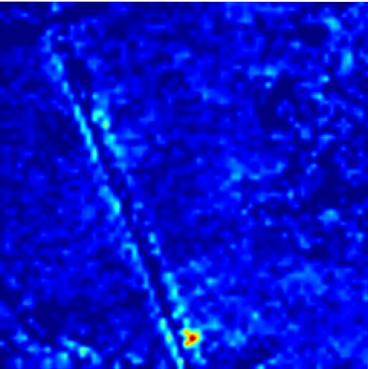}\vspace{2pt}
\end{minipage}}
\hspace{5pt}
\subfloat[$\bot\bot\bot\bot$]{
\begin{minipage}[t]{0.11\linewidth}
\includegraphics[width=1\linewidth]{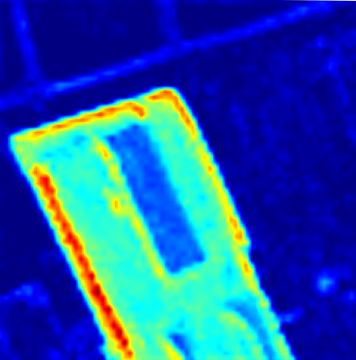}\vspace{2pt}
\includegraphics[width=1\linewidth]{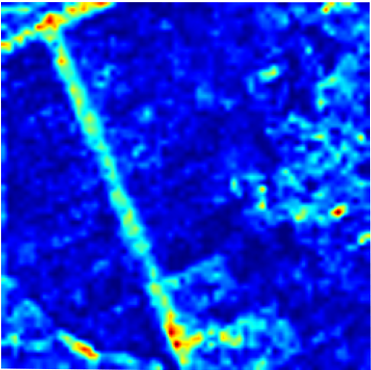}\vspace{2pt}
\end{minipage}}
\hspace{5pt}
\subfloat[$\bot\bot\circledcirc\circledcirc$]{
\begin{minipage}[t]{0.11\linewidth}
\includegraphics[width=1\linewidth]{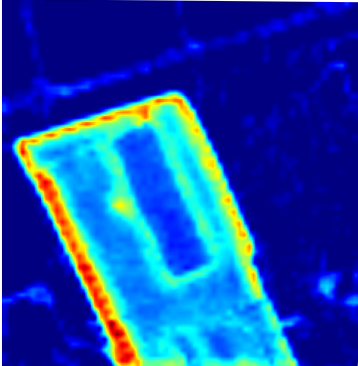}\vspace{2pt}
\includegraphics[width=1\linewidth]{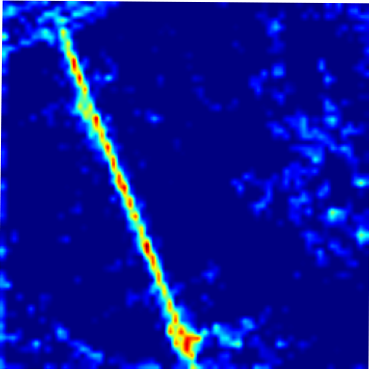}\vspace{2pt}
\end{minipage}}
\caption{Class activation maps generated by Grad-CAM for the change category of features modulated by the first stage (1/4 resolution of the input image) of the Feature Enhancer (FE). Example images are from the CLCD test set. The configurations represented by "$\circledcirc$" and "$\bot$" correspond to Table \ref{table:2}.}
\label{fig:clcd2}
\vspace*{-3mm}
\end{figure}

The proposed CDXLSTM achieves superior performance across three change detection datasets, outperforming recent methods, as shown in Tables~\ref{table:1}. This comprehensive comparison underscores the performance of our approach relative to leading techniques in the field. Specifically, CDXLSTM improves F1 scores by 0.38\%, 2.50\%, and 8.73\% on the LEVIR-CD, WHU-CD, and CLCD datasets, respectively, compared to ChangeMamba. This improvement is especially notable in complex scenarios like the CLCD dataset, which features diverse object distributions and richer variations. Furthermore, CDXLSTM is more efficient, with only 16.19M parameters and 3.92G Flops, significantly reducing computational cost compared to the latest methods AFCF3D-Net and RS-Mamba. 

In addition, visualization comparisons are shown in Fig. \ref{fig:clcd}.
CDXLSTM demonstrates more precise edge detection, whether for large objects (row 2) or small ones (row 3). Additionally, in multi-object detection scenarios, like in row 1, CDXLSTM excels by detecting finer details (especially at the bottom of the image) while maintaining clearer edge lines.

\subsection{Ablation study}

\textbf{Ablation of the long-term modeling strategy.}
We validate the effectiveness of XLSTM in long-term modeling within the feature extraction (FE) module, as shown in Table \ref{table:3}. We conduct experiments by replacing Bi-mLSTM with CNN (SCSEBlock in \cite{dtcdscn}), Transformer (Encoder in \cite{vaswani2017attention}), and Mamba (SS2D in \cite{liu2024vmamba}).  To demonstrate that Transformer performance still has limitations even after excluding quadratic complexity, we include linear Transformers AFFormer \cite{aff} (Linear-TR1) and FlattenSwin \cite{flatten} (Linear-TR2) in the comparison. We also introduce similar feature enhancement strategies, such as DBIM in HFIFNet \cite{han2025hfifnet}. The results show that Bi-mLSTM delivers the best performance, confirming XLSTM’s superior modeling capability over CNNs, Transformers, and Mambas.

\textbf{Ablation of the combination of Feature Enhancer.}
To explore the optimal combination of the Feature Enhancer layer (CTSR and CTGP) for extracting change representations from high- and low-resolution branches, we conduct relevant experiments including alternating between CTSR and CTGP, as shown in Table \ref{table:2}. Results indicate that CDXLSTM achieves the best performance with an F1-score of 78.73\% when using CTSR at 1/4 and 1/8 resolutions, and CTGP at 1/16 and 1/32 resolutions. As illustrated in Fig. \ref{fig:clcd2}, we further analyze Grad-CAM activation maps from the first stage of the Feature Enhancer to demonstrate the extraction of spatially refined features. Specifically, when only CTGP is used, the contours of the change regions are less distinct, and increased noise is observed in unchanged regions. When only CTSR is used, more false changes occur, and changes are not sufficiently refined. This highlights that our configuration not only reduces redundancy and noise but also leverages XLSTM's global perception capabilities while preserving essential locality.

\textbf{Ablation of the proposed blocks.}
We conduct an ablation study on the proposed modules CTSR, CTGP, and CSIF, as shown in Table \ref{table:4}. The results show that removing any of these modules leads to performance degradation. CTGP and CSIF, which focus on contours and large changes, yield more significant improvements in intuitive performance metrics compared to CTSR, which targets finer details like edges and small changes.

\section{Conclusion}

Mainstream RS-CD methods often struggle to balance performance and efficiency when relying on CNNs, transformers, and Mambas. However, CNNs are limited by their inability to effectively model global contexts, transformers are hindered by their quadratic computational complexity, and Mambas face restrictions due to their reliance on CUDA dependence and local correlation loss. To overcome these limitations, we propose CDXLSTM, which, for the first time, introduces XLSTM, offering linear complexity, global context awareness, parallel acceleration, and enhanced interpret-ability.
Specifically, we emphasize the importance of a scale-specific Feature Enhancer, combining CTSR and CTGP to better leverage XLSTM's global perception while preserving local details and reducing redundancy. Additionally, the CSIF module integrates global semantics from low-resolution change representations, maintaining spatial details. CDXLSTM achieves SOTA results on three RS-CD datasets, balancing accuracy and efficiency.

\bibliographystyle{IEEEtran}
\bibliography{myreferences.bib}

\vfill

\end{document}